# Applicability of Crisp and Fuzzy Logic in Intelligent Response Generation

T. V. Prasad

Dean, R &D and Industrial Consultancy
Lingaya's University
Faridabad, Haryana, India
tvprasad2002@yahoo.com

Sachin Lakra

Dept. of Information Technology
Manav Rachna College of Engg
Faridabad, Haryana, India
and Res. Scholar., K L University
sachinlakra@yahoo.co.in

G. Rama Krishna

Dept. of Comp. Sc. & Engg.
K L University
Vijayawada,
Andhra Pradesh, India
ramakrishna_10@yahoo.com

*Abstract*—**This paper discusses the merits and demerits of crisp logic and fuzzy logic with respect to their applicability in intelligent response generation by a human being and by a robot. Intelligent systems must have the capability of taking decisions that are "wise" and handle situations intelligently. A direct relationship exists between the level of perfection in handling a situation and the level of completeness of the available knowledge or information or data required to handle the situation. The paper concludes that the use of crisp logic with complete knowledge leads to perfection in handling situations whereas fuzzy logic can handle situations imperfectly only. However, in the light of availability of incomplete knowledge fuzzy theory is more effective but may be disadvantageous as compared to crisp logic.**

## I. INTRODUCTION

The generation of an intelligent response is evident in all human beings naturally. But whether intelligent responses can be generated by a robot or not is an issue under exploration presently. To some extent, artificial intelligence attempts to apply techniques that are similar to those used by human beings for solving various categories of problems that require intelligence. But these techniques have been only partially successful in achieving solutions at par with those of natural intelligence.

This paper explores the possibility of using soft computing techniques in combination with other artificial intelligence techniques for generating an intelligent response. A situation of giving an instruction to a robot is taken from the context of robotics. The applicability of each of the three soft computing techniques of neural networks, fuzzy logic and genetic algorithms and their combinations is considered and discussed.

The issue of the use of crisp logic as compared to fuzzy logic in general and the applicability of each in intelligent response generation is also discussed.

The outcome of the paper is that theoretically soft computing techniques can be used in intelligent response generation as they are nearer to human intelligence than hard computing techniques. On the other hand, the use of crisp logic and hard computing enables an intelligent system to complete tasks perfectly but the use of soft computing techniques does not lead to perfection. However, soft computing techniques also allow the handling of situations where available knowledge is incomplete. This is evident in cases where due to incomplete knowledge being available a hard computing based system is unable to take any action at all whereas a soft computing based system takes action to some extent which may actually be correct. It is again situation-dependent that no action being taken is found to be advantageous.

## II. SOFT COMPUTING TECHNIQUES

Computational techniques that fall under the purview of soft computing techniques include the following:
   a. Neural Networks
   b. Fuzzy Logic
   c. Genetic Algorithms

In these techniques, stringent, real-time or hard constraints, such as perfect accuracy, are not necessary. That is, the constraints under which decisions have to be computed are "soft" and hence the term soft computing. Examples of application of soft computing techniques are in speech-to-text conversion [1], solving the XOR problem [2] and filtering of speech signals [3].

## III. CRISP LOGIC VS FUZZY LOGIC

Crisp Logic is applicable where the knowledge about a situation is complete and completely certain. On the other hand, fuzzy logic is applicable where knowledge about a given situation is incomplete or minimal. In both the cases, the intelligent system has to generate a response which will ultimately lead to the completion of a given task.

In crisp logic, the knowledge available has to be hundred percent complete before a decision can be taken. The difficulty in this case is to ensure the acquisition of complete knowledge before the decision can be taken. Further, the intelligent system has to have the capability to predict that the use of specific steps will certainly lead to the completion of the task. This is possible if the steps to achieve the completion of the task are known completely and with certainty. Otherwise, in case of crisp logic if any of the steps are missing the intelligent system will not be able to complete the task.

On the other hand, if an intelligent system is unable to predict the completion of a task due to uncertainty in one or more steps it can still attempt to complete the task with whatever knowledge it has available, be it incomplete. This is possible with fuzzy logic only, which is a tool meant specifically to deterministically handle uncertainty in a robust manner.

## IV. APPLICABILITY OF SOFT COMPUTING IN INTELLIGENT RESPONSE GENERATION

### A. An Example in the Context of Robotics

Given a situation in which a robot or an intelligent system needs to carry out instructions given to it in the form of English sentences such as "It is very cool here in the room". For the system/robot to take "intelligent" action in this case, the following steps must be carried out:
   a. The robot/system has to interpret all the sentences that have been spoken by a human.
   b. The given sentence would lead to generation of the pre-requisites for any intelligent search space, i.e. initial state, goal state, set of rules, assumptions, etc.
   c. Next, in order to define a control strategy, the rules will have to be applied in a different manner (internally in the system, without causing any effect). This leads to generation of a huge search space.
   d. The most probable derivations/conclusions that can be drawn from the given sentence are
      a. That the room is too cool. The robot/system will have to check the temperature of the room using sensors. If the temperature is too low it will have to be raised. This could be accomplished as follows:
         (i) first locate a fan, if there is any, and then switch it off;
         (ii) locate an air conditioner, if there is any, and then switch it off;
         (iii) check if windows and/or doors are open, and then shut them properly;
         (iv) in the event, there is no air conditioner or fan functioning and the doors and windows are shut, then most likely it is winter season, and hence, heater should be located and switched on till the temperature rises to normal (i.e. around 25° Celsius)
      b. That the room is too cool which is identified by the use of sensors (so let it be, in comparison to the environment outside the room – so, no action required)

### B. Applicability of Soft Computing Techniques in the Example

The following soft computing techniques can be applied at various steps in the example above:
   a. Neural Networks to learn the knowledge of the temperature of the room gathered from the room using sensors. Neural networks can be used in both a crisp as well as a fuzzy manner.
   b. Reasoning and inferencing to identify the sources of the facts and/or rules isolated after visual and/or acoustic observation using sensors. Both reasoning and inferencing can be either crisp or fuzzy.
   c. Planning to identify the steps of the process to achieve the task of lowering the temperature and genetic algorithms to incrementally improve the solution in case the premises of the facts and/or rules are found to be incorrect or incomplete. In case the steps identified are incomplete the robot can identify the steps from its memory about any such task executed previously or will handle the task in a more efficient manner by using fuzzy decision making where the steps are partially known.
   d. In case the steps identified in step c are complete, the task can be completed in a crisp manner whereas if uncertainty exists about one or more of the steps the task can be handled in a less perfect but an effective manner using fuzzy logic.

### C. Applicability of Neural Networks in Intelligent Response Generation

Neural Networks can be used in either a crisp manner as standalone networks or in combination with fuzzy rules as neuro-fuzzy systems.

In case crisp values are received as input through the sensors in the example above, they can be input to a neural network, which will learn the values by adjusting the weights learned previously during its training, and based on the inputs it will output a crisp value if it has been so trained. On the other hand, the output desired in the above example is to interpret the temperature value as a linguistic variable named "low". This is possible only if fuzzy rules are incorporated into the neural network to create a neuro-fuzzy system. The result of the neural computation or the neuro-fuzzy computation will lead to an output that can be used for further computation in the later steps.

At this stage neuro-fuzzy computation is nearer to the way natural intelligence works in most human beings. But it leads to a fuzzy decision and an inadequate handling of a situation in many cases if the information or knowledge required to handle that situation are incomplete. On the other hand, if the information is complete and the decision is crisp the situation can be handled completely and perfectly.

### D. Applicability of Fuzzy Logic in Intelligent Response Generation

The term fuzzy implies imprecision. Thus the term fuzzy logic implies a logic that can handle imprecision. The root cause of the existence of imprecision is non-availability of either or all of complete knowledge or complete information or complete data. But human beings do handle most of the situations they encounter with such incompleteness and imprecision, in a robust manner. However, perfection can not be attained in such cases.

If a robot were to be given perfect knowledge about a given situation it will handle it perfectly and precisely because it will handle it in a crisp manner. This is what makes a robot superior to a human being. However, the robot will not appear to be natural in its behaviour and not even in its physical movements without the existence of fuzziness in its knowledge and its decision making and response generation capabilities.

Natural Intelligence has both fuzzy as well as crisp capabilities. It can handle both hard as well as soft computing almost equally well. But the capability to handle hard computing is far more prominent in a robot. The need today is to make robots capable of handling soft computing also, at least as well as human beings initially and then to go beyond human beings.

### E. Applicability of Genetic Algorithms in Intelligent Response Generation

Genetic algorithms are a mechanism by which improvements in a population or a set of data can be made in an incremental manner by eliminating the unfit members of a population or the unfit data. If the initial generation of the data obtained is insufficient or erroneous it can be improved over a few generations by using a genetic algorithm and can even be perfected. Perfect data can then be used in hard computing to achieve a task perfectly.

Perfect data can be used in a crisp manner to handle a task perfectly. On the other hand if the data is itself imprecise, it first has to be improvised through the use of a genetic algorithm and then applied to a situation. This may not be suitable in a hard computing environment due to the occurrence of a delay (considering present system efficiency) while the data is being improvised. If the data is to be improvised, the robot in the above example may have to obtain more knowledge for which it may or may not have the capabilities. For a robot to be able to carry out both hard as well as soft computing in a crisp manner it has to be capable of acquiring knowledge perfectly.

## V. CONCLUSION

The example cited in the paper leads us to conclude that both fuzzy as well as crisp techniques have their own areas of applicability when a given situation is to be handled and a given task is to be completed. The decision regarding whether to take a crisp approach or a fuzzy approach depends entirely upon the quantum of uncertainty about the steps involved in the completion of the task. The steps are certain if the knowledge, the information and the data required to handle the task is complete. However, acquiring perfect knowledge is an enormous challenge for any intelligent system. If the intelligent system is unable to acquire perfect knowledge about some of the steps it has to fill in the gaps using its intelligence. The brains of future robots will be imparted the ability to acquire perfect knowledge as well as the ability to fill in gaps, which will make them capable of achieving tasks perfectly and completely.

| S.No. | Steps of solving the problem | Soft Computing Technique applicable |
|---|---|---|
| a. | Interpret the sentence | Natural Language Processing + Fuzzy Logic |
| b. | Generation of the pre-requisites for any intelligent search space | Natural Language Processing + Fuzzy Logic |
| c. | In order to define a control strategy, apply the rules. This leads to generation of a huge search space. | Linguistic/Fuzzy rules |
| d. | Most probable derivations/conclusions from the given sentence are | Fuzzy/Crisp Reasoning and Inferencing |
| a. | That the room is too cool. If the temperature is too low it will have to be raised. This could be accomplished as follows: | Sensor Networks+Neural Networks |
| (i) | first locate a fan, if there is any, and then switch it off; | Pattern Recognition + Neural Networks |
| (ii) | locate an air conditioner, if there is any, and then switch it off; | Pattern Recognition + Neural Networks |
| (iii) | check if windows and/or doors are open, and then shut them properly; | Fuzzy/Crisp Motion + Planning + Genetic Algorithm |
| (iv) | If none of the above, most likely it is winter season, and hence, switch on heater till temperature is normal | Fuzzy/Crisp Motion + Planning + Genetic Algorithm |
| b. | That the room is less cool in comparison to the environment outside the room – no action required | No action required |